\documentclass[runningheads]{llncs}
\usepackage{graphicx}
\usepackage{comment}
\usepackage{amsmath,amssymb} 
\usepackage{color}
\usepackage{multirow}

\usepackage{amsmath}
\DeclareMathOperator*{\argmax}{arg\,max}

\begin{document}
\pagestyle{headings}
\mainmatter
\def\ECCVSubNumber{1147}  

\title{Mapping in a Cycle: Sinkhorn Regularized Unsupervised Learning for Point Cloud Shapes} 

\titlerunning{Mapping in a Cycle}
%
\author{Lei Yang\inst{1} \and
Wenxi Liu\inst{2,1} \and
Zhiming Cui\inst{1} \and
Nenglun Chen\inst{1} \and 
Wenping Wang\inst{1}
}
\authorrunning{L. Yang et al.}
%
\institute{
Department of Computer Science, The University of Hong Kong, China \\
\email{\{lyang, zmcui, nlchen, wenping\}@cs.hku.hk}
\and
College of Mathematics and Computer Science, Fuzhou University, China \\
\email{wenxi.liu@hotmail.com}
}

\maketitle

\begin{abstract}
We propose an unsupervised learning framework with the pretext task of finding dense correspondences between point cloud shapes from the same category based on the cycle-consistency formulation. In order to learn discriminative pointwise features from point cloud data, we incorporate in the formulation a regularization term based on Sinkhorn normalization to enhance the learned pointwise mappings to be as bijective as possible. Besides, a random rigid transform of the source shape is introduced to form a triplet cycle to improve the model's robustness against perturbations. Comprehensive experiments demonstrate that the learned pointwise features through our framework benefits various point cloud analysis tasks, e.g. partial shape registration and keypoint transfer. We also show that the learned pointwise features can be leveraged by supervised methods to improve the part segmentation performance with either the full training dataset or just a small portion of it.
\keywords{Point cloud, unsupervised learning, dense correspondence, cycle-consistency}
\end{abstract}

\section{Introduction}\label{sec:intro}

Point clouds are unordered sets of interacting points sampled from surface of objects for 3D shape representation, and have been widely used in computer vision, graphics, robotics, etc. for their accessibility and flexibility. 
With the recent advancement of deep learning techniques, a spectrum of networks have been proposed to process point cloud data and to learn to perform various tasks, e.g.~\cite{qi2017pointnet,qi2017pointnet++,li2018pointcnn,lee2018set,wang2019dynamic}, which have achieved tremendous progress. 
However, a major limitation of deep networks is their data hunger nature that requires a large amount of supervisory signals to learn a satisfactory task-specific model. 
Therefore, many attempts have been made to alleviate this issue, and among others training deep networks in an unsupervised manner (without manually labeled data) shows its potential in many scenarios~\cite{devlin2018bert,chen2020simple,misra2019self}.
In the case of 3D point clouds, these techniques are in demand as it is prohibitive to attain accurate, densely labeled ground-truth on point clouds for various shape analysis tasks. 

\begin{figure}[t!]
    \centering
    \includegraphics[width=0.8\textwidth]{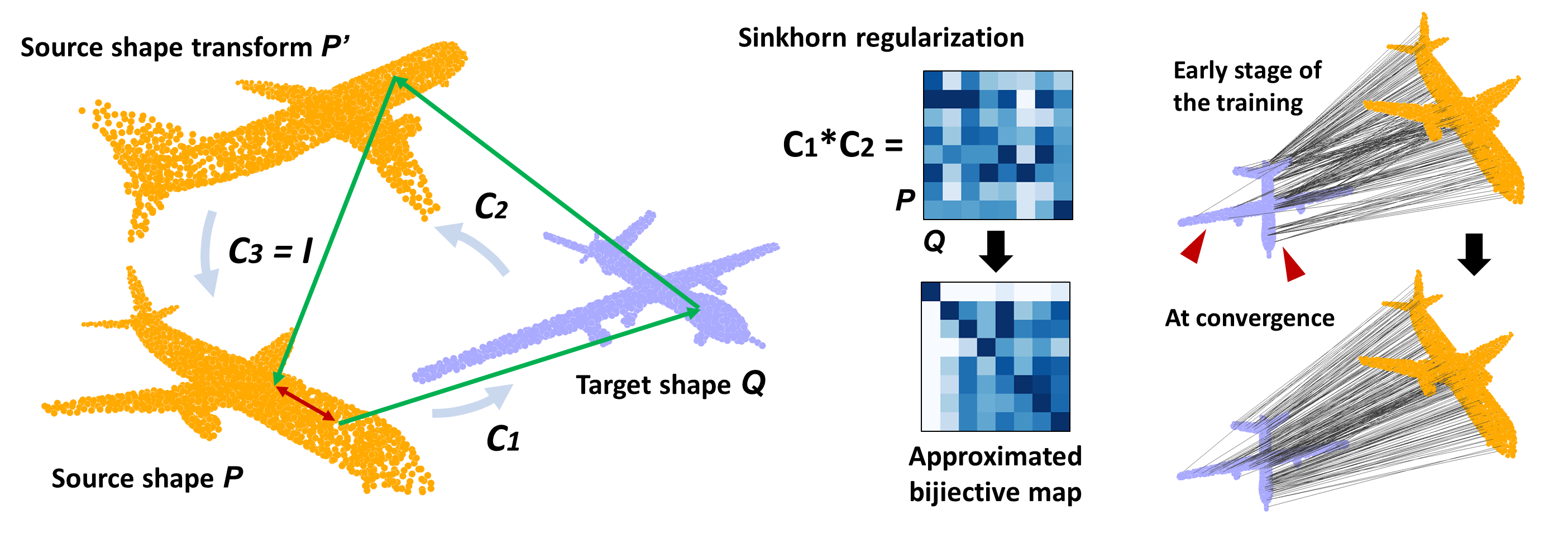}
    \caption{We train a neural network in an unsupervised manner to derive dense correspondences between point cloud shapes based on the cycle-consistency formulation. Given a source shape $\mathcal{P}$, its rigid transform $\mathcal{P}'$ and a target one $\mathcal{Q}$, the cycle is completed by mapping every point in $\mathcal{P}$, via $C_1$, $C_2$, $C_3$, and finally back to $\mathcal{P}$ (left). 
    The red segment indicates the measure of the cycle deviation which is minimized during the unsupervised training. Within the cycle, the correspondence map formed by the mappings $C_1$ and $C_2$ is constrained to approximate a bijiective map by Sinkhorn regularization, significantly reducing the number of many-to-one correspondences (right).}
    \label{fig:teaser}
\end{figure}

As one of unsupervised learning approaches, algorithms based on cycle consistency have attracted interests in many vision-based applications, e.g., video object segmentation~\cite{wang2019learning} and facial landmark detection~\cite{thewlis2017unsupervised}, as well as some recent 3D shape analysis works~\cite{huang2013consistent,zhou2016learning,wang2019prnet,groueix2019unsupervised}. 
Intuitively, a cycle consistent transformation between a pair of data instances shall map one data instance to the other, and then transform it backward with little deviation. With a pretext defined, such as registration~\cite{wang2019prnet} or deformation~\cite{groueix2019unsupervised}, one can leverage the cycle consistency formulation between a pair of unlabeled data, model the transformation with a neural network, and thus optimize the network parameters by minimizing the cycle deviation.

In this work, we leverage such a formulation to pre-train the neural network in an unsupervised manner, and aim to learn pointwise features for subsequent 3D point cloud applications. Specifically, the pretext in our setting is to find dense correspondences between two point cloud shapes using the learned pointwise features. In particular, given a pair of source and target shapes, we intend to find, for each point in the source, its corresponding point in the target. Then, starting from the target ones, we search reversely their corresponding points in the source. During this process we minimize the cycle deviation of each reversely corresponded point from its original location in the source. In this way, the network parameters can be optimized and expected to encode each point to a high-dimensional descriptor for correspondence query.

While dense correspondences between contents have been exploited in many image-based applications (e.g.~\cite{wang2019learning,thewlis2019unsupervised}), this self-supervised framework encounters two major challenges when dealing with point cloud data. 
First, since the point clouds are usually sampled from smooth 3D surfaces, each point embeds very limited information as opposite to image pixels with rich textures. 
This precludes the network training based on cycle consistency as the obtained correspondence may map a point to a wrong but spatially proximate location, forming a many-to-one mapping while yielding a small loss. 
Thus, the learned representation may suffer from this sub-optimality and fail to attain sufficient distinctiveness for correspondence query and potential applications.
Second, many point-based networks based on cycle consistency assume the shapes are well aligned, and thus are sensitive to rotations. This makes the extracted features unrobust to small perturbations, and may become less applicable in many applications. 

To address the first and primary concern, we propose a novel regularization technique to strengthen the pointwise correspondences to be as bijective as possible.  
We thus impose a bijective constraint on the cycle-back correspondence by adopting the Sinkhorn normalization technique~\cite{sinkhorn1964relationship,knight2008sinkhorn}. We term this constraint as Sinkhorn regularization in our paper. 

We also introduce, into the cycle, an additional shape which is a random rigid transform of the source shape, forming a 3-edge cycle as shown in Fig.~\ref{fig:teaser}.
In this particular setting, each point starting from the source first finds its corresponding point in the target (i.e. $C_1$ in Fig.~\ref{fig:teaser}), and then arrives at the source transform (i.e. $C_2$). 
Since the last transport edge (i.e. $C_3$) from the source transform to its origin provides us the ground-truth dense correspondences that form a one-to-one map, we can safely impose the bijective constraint by Sinkhorn regularization on this particular edge without assuming any shape pair should meet the bijective constraint. 
Further, unlike traditional cycle consistency methods on shapes, the introduction of a transformed shape allows the network and the learned pointwise features to be less sensitive against non-aligned point cloud data. 
This partially addresses the second challenge as mentioned before, thus making the learned pointwise features appealing to many downstream tasks. 

To demonstrate the effectiveness of our proposed framework, we conduct comprehensive experiments, in which we leverage the pointwise features learned from our model to perform partial shape registration, keypoint transfer, and as an additional pre-trained feature for supervised part segmentation. 
In these experiments, it is demonstrated that our approach can surpass the state-of-the-art methods or be comparable to their performances. Contributions of this paper are summarized as follows: 

1) A novel unsupervised learning framework based on cycle-consistent dense correspondences between point cloud shapes of the same category;

2) The Sinkhorn regularization that brings a large improvement on learning discriminative pointwise features;

3) Extensive evaluation showing the effectiveness of the proposed unsupervised learning strategy in multiple point cloud analysis tasks.

\section{Related Work}

\textbf{Deep unsupervised methodology and applications.}
Unsupervised learning methodology has emerged to address one of the major concerns for data-driven algorithms – the need for a large set of labeled data, and has achieved state-of-the-art performances in many applications in language~\cite{devlin2018bert} and images~\cite{chen2020simple,he2019momentum}. To achieve this, a pretext is often required for network pretraining and representation learning, such as by contrastive learning~\cite{oord2018representation,wu2018unsupervised,misra2019self,he2019momentum}, mutual information maximization~\cite{tschannen2019mutual}, or via reconstruction~\cite{sauder2019self} and correspondence~\cite{thewlis2017unsupervised,dwibedi2019temporal,wang2019learning}.

\textbf{Deep unsupervised point cloud analysis.}
Point-based networks have demonstrated their capability in many shape analysis tasks, such as classification, segmentation, reconstruction and registration~\cite{qi2017pointnet,qi2017pointnet++,wu2019pointconv,li2018pointcnn,wang2019dynamic,deprelle2019learning,zeng20173dmatch}.
To enforce the network to learn semantically meaningful and consistent representations, many pretext tasks have been designed such as part re-arranging~\cite{sauder2019self}, half-to-half prediction~\cite{han2019multi}, deformation~\cite{groueix2018atlasnet,groueix2019unsupervised}, and
self-supervised classification and clustering~\cite{hassani2019unsupervised}. Many of these works rely on the reconstruction metric as an indicator for the unsupervised training process~\cite{groueix2018atlasnet,zhao20193d,deprelle2019learning}. In this paper, we provide an alternative viewpoint that is complementary to the prior works. Instead of geometric reconstruction of the content, we consider the pretext of finding dense correspondences between shapes, and solve it as a \textit{soft permutation recovery problem} for the indices of points in point cloud.

\textbf{Learning from 3D shape correspondences.}
3D shape correspondence has long been an exciting topic in 3D shape modeling~\cite{van2011survey,kim2012exploring,kim2013learning,huang2013fine,sahilliouglu2019recent,halimi2019unsupervised}. Many state-of-the-art works have leveraged dense correspondences to learn and perform various shape analysis tasks. 
\cite{deng2018ppf,deng2018ppfnet,zeng20173dmatch,choy2019fully} design network architectures to learn local contextual descriptors via the task of 3D scan registration. This is amiable especially in the case of scene understanding and processing. As for the analysis of man-made shapes, \cite{wang2019deep,wang2019prnet} and \cite{aoki2019pointnetlk} instill classical methodologies (e.g. iterative closest point) in the neural network design and achieve state-of-the-art performance. In our case, we are more focused on learning pointwise features that are consistent across man-made shapes and thus differ from these studies.

In pursuit of such pointwise latent representations of man-made shapes, 
\cite{huang2017learning,chen2019edgenet,muralikrishnan2019shape} make use of rough dense shape correspondence as supervision and demonstrate promising performance in shape correspondence.
Alternatively, cycle consistency, initially proposed in \cite{huang2013consistent}, has been widely employed to establish correspondences between diverse data in an unsupervised manner, on images~\cite{CycleGAN2017,zhou2016learning}, videos~\cite{wang2019learning,dwibedi2019temporal}, and more recently on point cloud data~\cite{groueix2019unsupervised}.

In line of these prior works, we build our unsupervised learning framework based on cycle consistency to process point cloud shapes. Different from prior arts~\cite{groueix2019unsupervised}\cite{wang2019prnet} that evaluate cycle consistency by measuring shape discrepancy, we innovatively cast the problem of finding dense correspondences as solving permutation on point clouds.
This particular design provides an alternative view to existing works and allows the network to learn a pointwise representation. While the unsupervised learning works ~\cite{groueix2019unsupervised,wang2019prnet} focus on their pretexts such as deformation and registration, we show a variety of applications with the proposed network as well as the learned pointwise representation. We further propose a novel Sinkhorn regularization in the cycle consistency framework to enforce the learned pointwise features to be sparsely corresponded for different instances. 
\section{Methodology}

Our overarching goal is to learn, without manual labels, a category-specific pointwise encoding that benefits the downstream applications such as shape registration, keypoint detection and part segmentation. 
To this end, we train the network via a pretext task of finding dense correspondences between two point cloud shape instances based on the cycle-consistent pointwise encoding features learned through the proposed framework. 

\subsection{Unsupervised loss based on cycle consistency}
\label{sec:cycle_loss}

The pretext of finding cycle-consistent dense correspondences is depicted in Fig.~\ref{fig:teaser}.
We denote the source shape $\mathcal{P}$, its random rigid transform $\mathcal{P}'$, and the target $\mathcal{Q}$, forming a 3-edge cycle from $\mathcal{P}$ to $\mathcal{Q}$ (mapping $C_1$), then from $\mathcal{Q}$ to $\mathcal{P}'$ (mapping $C_2$), and finally return to $\mathcal{P}$ from $\mathcal{P}'$ (mapping $C_3$). In order to formulate it as an optimization problem, with the dense correspondences between shapes, our goal is to minimize the deviation between each cycle-back point and its origin (i.e. the red segment in Fig.~\ref{fig:teaser}), thus enforcing the cycle consistency.

\textbf{Correspondence query between point cloud shapes. }
We use a point-based neural network, denoted as $f_{\theta}$ with trainable parameters $\theta$, to learn the pointwise features, which will be employed for the correspondence query.  
We denote a point using its index as $p_k$ where $p_k \in \{0,1\}^{|\mathcal{P}|}$ is a one-hot vector with the $k$-th entry equal to $1$ and the rest to $0$. The corresponding 3D coordinate of $p_k$ is denoted by $\mathbf{p}_k$.
If a particular point $q_i$ from shape $\mathcal{Q}$ is said to correspond to $p_k$ from $\mathcal{P}$, then the associated learned representation $f_{\theta}(q_i)$ is more similar with $f_{\theta}(p_{k})$ than all other points in $\mathcal{Q}$ as below:
\begin{equation}\label{eq:correspondence}
    i = \argmax_{q_j \in \mathcal{Q}} S(f_{\theta}(p_k), f_{\theta}(q_j)),
\end{equation}
where $S(\cdot,\cdot)$ measures the similarity between any two pointwise representations and is defined as their inner product.
Since the operator $\argmax$ is not differentiable, we approximate the solution to the above equation by a scaled softmax function:
\begin{equation}\label{eq:softmax_approx}
    q_i \approx  C(\mathcal{Q}, p_k; f_{\theta})
    = \frac{\exp(f_{\theta}(p_k)^Tf_{\theta}(q_j)/\tau)}{\sum_j\exp(f_{\theta}(p_k)^Tf_{\theta}(q_j)/\tau)},
\end{equation}
where $C(\mathcal{Q}, p_k; f_{\theta})$ is a vector that represents the probability that $p_k$ corresponds to all points in $\mathcal{Q}$.
Thus, the dense correspondences from $\mathcal{P}$ to $\mathcal{Q}$ can be approximated as follow:
\begin{equation}\label{eq:transport}
    \mathbf{Q}  \approx C(\mathcal{Q}, \mathcal{P}; f_{\theta}).
\end{equation}
where, ideally, $\mathbf{Q}$ is expected to be a permutation matrix, establishing a one-to-one mapping between two given shapes $\mathcal{P}$ and $\mathcal{Q}$.

\textbf{Cycle-consistency loss. }
In this paper, we use three shapes to form a 3-edge cycle $\{\mathcal{P} \rightarrow \mathcal{Q} \rightarrow \mathcal{P}' \rightarrow \mathcal{P}\}$, where $\mathcal{P}$ and $\mathcal{Q}$ are termed the source and the target shapes, respectively, and $\mathcal{P}'$ is a random rigid transform of $\mathcal{P}$ that helps increase the robustness of the model.
Thus, with the cycle-consistency condition met, this closed cycle should finally bring every point (in terms of index and not spatial coordinates) back to its origin index via the following mappings,
\begin{equation}\label{eq:cycle_consistency_condition}
    C_{cycle}(\mathcal{P}) = C_3(\mathcal{P},\mathcal{P}')C_2(\mathcal{P}',\mathcal{Q})C_1(\mathcal{Q},\mathcal{P})
    =C_3(\mathcal{P},\mathcal{P}')C_{1,2}(\mathcal{P}',\mathcal{P}),
\end{equation}
where $C_{cycle}(\mathcal{P})$ shall be the identity matrix that brings points in $\mathcal{P}$ back to its origin index via $C_1$, $C_2$, and $C_3$. 
Similarly, $C_{1,2}(\mathcal{P}',\mathcal{P})$ forms the mapping from the source shape $\mathcal{P}$ to the transformed shape $\mathcal{P}'$ via $\mathcal{Q}$. 
To measure the cycle deviation from the above formulation, a loss should be defined
\begin{equation}\label{eq:deviation}
    d_{cycle} = D(\mathbf{I}_{|\mathcal{P}|}, C_{cycle}(\mathcal{P})),
\end{equation}
where $\mathbf{I}_{|\mathcal{P}|}$ is the identity matrix of size $|\mathcal{P}|$.

As we introduce a rigid transform to the end of the cycle list, the cycle mapping mainly depends on two parts, i.e., $C_{1,2}(\mathcal{P}',\mathcal{P})$ and $C_3(\mathcal{P},\mathcal{P}')$, in Eq.~\ref{eq:cycle_consistency_condition}. 
First, as rigid transformations in $\mathbb{R}^3$ do not alter the permutation of the point cloud data. So, when it is perfectly estimated, $C_3(\mathcal{P},\mathcal{P}')$ should be the identity matrix that maintains the original permutation of $\mathcal{P}$.
On the other hand, the mapping $C_{1,2}$ from $\mathcal{P}$ to $\mathcal{P}'$ (via $\mathcal{Q}$), in an ideal situation, should be the identity matrix as well. 
Hence, the cycle loss minimization can be reduced to minimize two terms, $D(\mathbf{I}_{|\mathcal{P}|}, C_{1,2}(\mathcal{P}',\mathcal{P}))$ and $D(\mathbf{I}_{|\mathcal{P}|}, C_3(\mathcal{P},\mathcal{P}'))$.

One way to concretely define $D(\cdot, \cdot)$ is to use KL-divergence or cross-entropy losses to formulate the problem as classification. 
However, minimizing such \textit{thousand}-way classification losses may be difficult at the beginning and overlook, during the course of optimization, the underlying geometric relationship of the points cloud shapes. Therefore, we cast the cycle consistency loss in a regression form similar to the losses used in~\cite{dwibedi2019temporal,thewlis2019unsupervised}. This way, we impose a soft penalty on the wrong cycles relying on the distances from their correct correspondences,
\begin{equation}\label{eq:regression_cycle_loss}
    \mathcal{L}_{C} = \| D(\mathcal{P}) \otimes C_{1,2}(\mathcal{P}', \mathcal{P})\|_1,
\end{equation}
where $D(\mathcal{P}) = \{d_{p,p'} = d_{Euclid}(\mathbf{p}, \mathbf{p}'), \forall \mathbf{p}, \mathbf{p}' \in \mathcal{P}\}$ measures the Euclidean distance between a pair of points in $\mathcal{P}$ and $\otimes$ is the element-wise product. Here the Euclidean distance is adopted for simplicity and computational efficiency, but one may employ more accurate geodesics distance for training. Note that the diagonals of $D(\mathcal{P})$ are zeros, which makes the loss of Eq.~\ref{eq:regression_cycle_loss} to be zero when $C_{1,2}(\mathcal{P}',\mathcal{P})$ converges to be an identity matrix. This loss is thus equivalent to the classification-based formulation at convergence, while additionally taking the spatial proximity of points into consideration. 
Similarly, we formulate $D(\mathbf{I}_{|\mathcal{P}|}, C_{3}(\mathcal{P},\mathcal{P}'))$ as follow:
\begin{equation}\label{eq:robust_rotation}
    \mathcal{L}_{R}= \|D(\mathcal{P}) \otimes C_{3}(\mathcal{P},\mathcal{P}')\|_1.
\end{equation}

\subsection{Sinkhorn regularization for bijective constraint}
\label{sec:bijective_constraint}

Optimizing the regression-based cycle loss can converge to the correct correspondences as demonstrated in many image-based applications~\cite{dwibedi2019temporal,thewlis2019unsupervised}. However, the convergence will be slowed down or even get stuck in the case of 3D point cloud data. This is because the decaying distance-based penalty imposed in Eq.~\ref{eq:regression_cycle_loss} cannot provide a sufficient magnitude of loss that encourages the network to distinguish nearby points as the optimization proceeds. Thus, it may still result in many-to-one mappings and thus wrong cycles, leading to undesirable results.

To address this issue, we introduce a so-called Sinkhorn regularization term, $L_S$, in addition to the previous ones. This design relies on the fact that $C_{1,2}(\mathcal{P}',\mathcal{P})$ in our setting should ideally form a bijective map.
Instead of directly enforcing $C_{1,2}$ to be the identity matrix, we \textit{relax} this constraint to any permutation matrices, retaining the bijective property.
The reason of using a relaxed bijective map instead of the identity is that while this relaxation penalizes the deviation of $C_{1,2}$ from a permutation, the synergistic effect of $L_S$ and $L_C$ gradually makes $C_{1,2}$ converge to the identity as the training proceeds. 
This novel relaxation brings the performance gain by a large margin in terms of the percentage of correct cycle-consistent correspondences, as shown in the ablation study (Sec.~\ref{sec:ablation}).

We follow the methods proposed in~\cite{adams2011ranking,mena2018learning} to enforce this constraint and describe it to make our paper self-contained. 
One may compute the optimal approximant to $C_{1,2}$ from the permutation set $\mathbb{P}$ with dimension $|\mathcal{P}|$,
\begin{equation}\label{eq:perm_argmax}
    {X^*} = \argmax_{{X} \in \mathbb{P}_{|\mathcal{P}|}} \langle {X}, {C_{1,2}} \rangle_F,
\end{equation}
where $\langle X, C_{1,2} \rangle_F$ denotes the Frobenius inner product of the two matrices. As solving the above linear assignment problem (Eq.~\ref{eq:perm_argmax}) is generally NP-hard, the constraint can be further relaxed to solve the best approximation of $C_{1,2}$ from the set of doubly stochastic matrices $\mathbb{B}_{|\mathcal{P}|}$,

\begin{equation}\label{eq:birkhoff_apprx}
    \Tilde{X} = \argmax_{X \in \mathbb{B}_{|\mathcal{P}|}} \langle X, C_{1,2} \rangle_F.
\end{equation}
Solving the maximization problem of Eq.~\ref{eq:birkhoff_apprx} has been shown to be exceptionally simple by taking row-wise and column-wise softmax normalizations in an alternating fashion. This is known as the Sinkhorn normalization~\cite{sinkhorn1964relationship} where $\tilde{X}$ shall meet the following conditions at convergence: 
\begin{equation}
    \Tilde{X}\mathbf{1} = \mathbf{1}, \quad \Tilde{X}^T\mathbf{1} = \mathbf{1}, \quad \text{and } \Tilde{X} \in \mathbb{B}_{|\mathcal{P}|}. \nonumber
\end{equation}

While the solution to Eq.~\ref{eq:birkhoff_apprx} can be reached in a limit sense, practically a truncated Sinkhorn normalization~\cite{adams2011ranking} is used to obtain a fair approximation,
\begin{equation}
    \tilde{X} \approx SH(C_{1,2}; t, l),
\end{equation}
where two hyper-parameters, i.e., the number of iterations $l$ for the column-wise and row-wise normalization and the temperature $t$ for softmax normalization are to be furnished. 
We adopt this truncated Sinkhorn normalizaltion and set $t$ and $l$ to be 0.3 and 30 across all our experiments.

\textbf{Sinkhorn regularization. }
Accordingly, during the network optimization we add the following Sinkhorn regularizaton to the loss function:
\begin{equation}
    \mathcal{L}_{S} = \| C_{1,2}(\mathcal{P}', \mathcal{P}) - SH(C_{1,2}; t,l) \|_{1}.
    \label{eq:ls}
\end{equation}
This loss term enforces $C_{1,2}$ to be a bijective map, and thus encourages the neural network $f$ to learn discriminative pointwise features by reducing many-to-one mappings between point clouds $\mathcal{P}$ and $\mathcal{Q}$ which $C_{1,2}$ traverses. 
As it is derived based on $C_{1,2}$, $SH(C_{1,2}; t, l)$ keeps its closeness to $C_{1,2}$ at every iteration step. Thus, this formulation provides a gradual guidance for $C_{1,2}$ to become a permutation matrix ensuring the bijective property.

\subsection{Loss function}

To sum up, the loss function of our unsupervised framework consists of three terms, as below:
\begin{equation}\label{eq:full_loss}
    \mathcal{L} = \lambda_{C} \mathcal{L}_{C} + \lambda_{R} \mathcal{L}_{R} + \lambda_{S} \mathcal{L}_{S},
\end{equation}
where $\lambda_{C}$, $\lambda_{R}$, and $\lambda_{S}$ are predefined coefficients for balancing these loss terms.
By the loss term $\mathcal{L}_C$, we can constrain the chained mapping $C_{1,2}(\mathcal{P}', \mathcal{P})$ to be an identity matrix. 
In addition, as we explicitly require the last shape to be some random rigid transform $\mathcal{P}'$ of the source $\mathcal{P}$, 
$L_R$ enforces the learned representation to be robust to mild rotations.
Moreover, $L_S$ encourages the correspondence to be as bijective as possible, benefiting the pointwise representation learning.

\subsection{Network architecture}
\label{sec:network_arch}

Our network takes 1024 points as input from a point cloud shape sampled by the farthest sampling technique, and outputs pointwise representations of dimension 64.
PointNet++~\cite{qi2017pointnet++} is adopted as the backbone for unsupervised learning, and trained with Eq.~\ref{eq:full_loss}. Other backbones (e.g. DGCNN~\cite{wang2019dynamic}) may be applicable but we limit our discussion here to PointNet++.

In addition, we incorporate the multi-head attention module (see~\cite{vaswani2017attention,lee2018set}) in between adjacent layers of the original PointNet++.
The motivation of such design is to combine both local information gathered by convolution-like operations and non-local information by self-attention modules to benefit the unsupervised learning process. For details of the network architecture, please refer to our supplementary materials.

\section{Experimental Results}
\label{sec:exp}

\subsection{Implementation and results of the unsupervised pretraining}

We pre-train the proposed networks on the pretext of finding dense correspondences on \textit{ShapeNet part segmentation} dataset. 
\textit{ShapeNet part segmentation} dataset contains 16 categories of shapes, each of which is represented by a point set and associated with normals and the part labels. 
The number of parts per shape in a category varies from 2 to 6, totaling 50 parts in the dataset. All the point clouds are pre-aligned. 

Training data are augmented with random rotations, translations and scalings sampled from a uniform distribution. 
The rotations and translations along each coordinate axis are sampled from the range $[-15 deg, +15 deg]$ and $[-0.2, +0.2]$, respectively; and the uniform scaling factor ranges from $[0.8, 1.25]$. 
These random transformations are applied to each of the training triplet, i.e., source shape, its rigid transform, and the target.
We first sample a source shape and a target one from the same category, and then apply two sets of random transformations described above to the source and one set of random transformations to the target, respectively. 
Thus, three transformed shapes are generated, forming a triplet, i.e. the source, its rigid transform, and the target, for training.

We use a variant~\cite{reddi2019convergence} of the Adam optimizer~\cite{kingma2014adam} with $\beta_1=0.900$ and $\beta_2=0.999$ across all experiments presented in the paper. 
The learning rates for bias and the rest of parameters are set to $0.0001$ and $0.0005$, respectively, without decay during the training process. 
Balancing coefficients $\lambda_{C,R,S}$ are set to $1.0,1.0,0.06$ to weight each loss term for the network pre-training.
All network models are trained with an NVIDIA GeForce GTX 1080Ti graphical card.

\begin{figure}[t]
    \centering
    \includegraphics[width=0.9\textwidth]{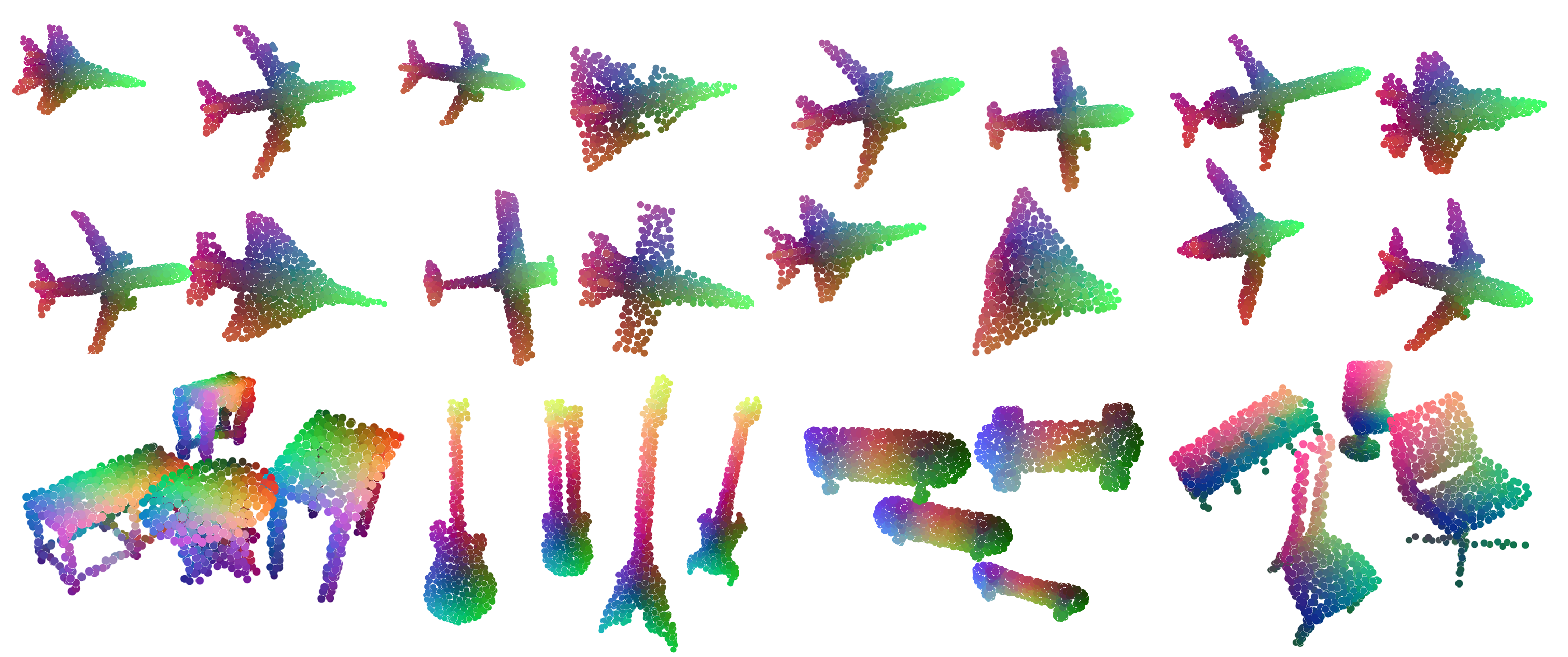}
    \caption{Visualization of the learned pointwise representations on rotated shapes from different categories, i.e., \textit{Airplane}, \textit{Table}, \textit{Guitar}, \textit{Skateboard}, and \textit{Chair}. 
    Color-codes reflect the consistency of the learned pointwise representations across a variety of shape instances, even if the shapes are under perturbation of rigid transformations}
    \label{fig:correspondence_results}
\end{figure}

We randomly sample a set of perturbed shapes with random rigid transformations as described above, and visualize their pointwise features in Fig.~\ref{fig:correspondence_results}. 
The features are dimension-reduced via t-SNE~\cite{maaten2008visualizing} and color-coded to reflect their similarity.
Although the shapes are randomly transformed, the visualization shows that our learned representations on various non-aligned shape instances are consistent. 
More qualitative results can be found in the supplementary.

\subsection{Ablation study}\label{sec:ablation}

We validate our network design and our proposed training strategy in this subsection. 
For evaluation, we employ the ratio of correct cycle matches ($CC\%$) during training and validation as the metrics, which indicates the success rate of completing cycle-consistent correspondences. 
Point-cloud shapes from three categories, i.e. Airplane, Chair, and Table, from the \textit{ShapeNet part segmentation} dataset are adopted for evaluation.

Different settings are compared to justify our proposed framework with the designed loss function (Eq.~\ref{eq:full_loss}) and the self-attention module.
We first evaluate two variants of the loss functions: 1) removing the bijective constraint enabled by the Sinkhorn regularization (\textit{w/o $L_{S}$}); and 2) enforcing the correspondence matrix to be the identity matrix instead of permutation matrices (i.e. replacing Eq.~\ref{eq:ls} by $L_{I} = \| C_{1,2} - \textbf{I}\|_1$, denoted \textit{$L_I$}). 
The comparison results are depicted in Tab.~\ref{tab:ablation}. 
As can be seen from the second row block of Tab.~\ref{tab:ablation}, our complete loss function (\textit{Ours}) produces the best result among the three settings. 
The performances of the other two settings, i.e. \textit{w/o $L_{S}$} and \textit{$L_I$}, are similar.

The performance gain by Sinkhorn regularization is mainly due to the penalty it imposes on the many-to-one correspondences, which smoothly increases as the training proceeds and thus drives the resulting mapping as much close to a bijective map as possible.
On the contrary, the setting without such a constraint (\textit{w/o $L_{S}$}) would indulge many-to-one mappings; and the setting (\textit{$L_{I}$}) that enforces the mapping to identity would be too difficult for training at the very beginning stage, thus impeding the convergence. 

\begin{figure}
    \centering
    \includegraphics[width=\textwidth]{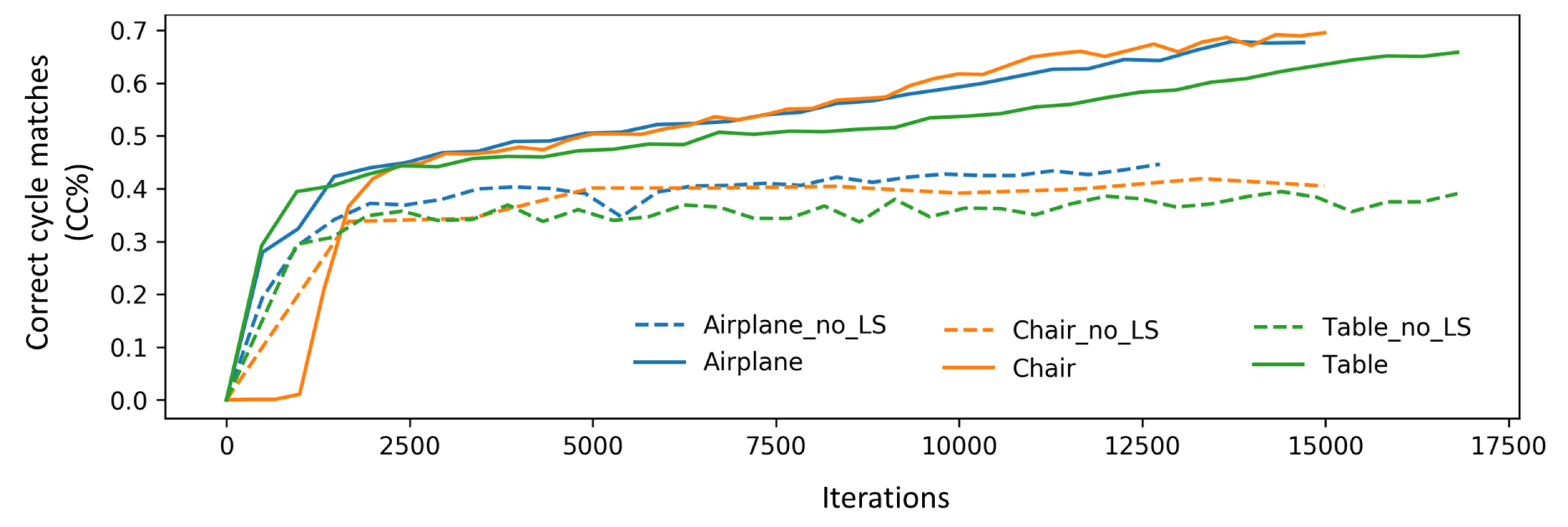}
    \caption{Performances with the Sinkhorn regularization $L_S$ (solid) or without it (dashed) are compared in terms of the percentages of correct cycle matches (CC$\%$) on three data categories, showing that the Sinkhorn regularization can facilitate the optimization and achieves consistently better results}
    \label{fig:ablation_curve}
\end{figure}

As shown in Fig.~\ref{fig:ablation_curve}, we compare the $CC\%$ of testing using the models obtained at different training iterations, in which the higher results are better. 
As observed, after training for more than 2000 iterations, the results w/ $L_{S}$ (in solid curves) perform significantly better than the ones w/o $L_{S}$ (in dashed curves), which shows the advantage of our proposed Sinkhorn regularization.

In addition, we compare our network structure with self-attention modules against the vanilla PointNet++. 
As observed from Tab. \ref{tab:ablation}, our results are higher than those produced by the vanilla PointNet++. 
This is primarily because the self-attention modules will attend to long-range dependency that convolution-like operations overlook at the entrancing levels.
Note that, although using the network structure equipped with self-attention modules, the two settings (\textit{$L_{I}$} and \textit{w/o $L_{S}$}) are generally inferior to the vanilla PointNet++ (\textit{w/o self-attention}) trained with the Sinkhorn regularization, revealing that it is the primary contributor to the performance gain. 
Besides, we also evaluate the input of our model. As shown in Tab.~\ref{tab:ablation}, the performance will be degraded without normals as inputs. 
But such a decrease in performance is relatively small, comparing to settings of \textit{$L_{I}$} and \textit{w/o $L_{S}$} that use normals.

\setlength{\tabcolsep}{4pt}
\begin{table}[t]
\begin{center}
\caption{Ablation study on different loss terms, network structures, and the input of our model. $CC\%$ denotes the percentage of the correct cycle matches. We compare the metrics on three categories of data: \textit{Airplane}, \textit{Chair} and \textit{Table}}
\label{tab:ablation}
\setlength{\tabcolsep}{10pt}
\scalebox{0.8}{
    \begin{tabular}{l|| l l | l l | l l | l l}
    \hline\noalign{\smallskip}
    Category & \multicolumn{2}{c|}{Airplane} & \multicolumn{2}{c|}{Chair} & \multicolumn{2}{c|}{Table} & \multicolumn{2}{c}{Mean}\\
     $CC\%$ & Train & Val. & Train & Val. & Train & Val. & Train & Val.\\
    \noalign{\smallskip}
    \hline
    \noalign{\smallskip}
    \textbf{Ours} & 69.4 & 67.9 & 68.7 & 69.4 & 67.1 & 65.1 & 68.4 & 67.5 \\ 
    \noalign{\smallskip}
    \hline
    \noalign{\smallskip}
    w/o $L_{S}$ & 44.8 & 44.9 & 40.4 & 41.0 & 40.5 & 39.1 & 41.9 & 41.5 \\
    $L_{I}$ (replacing $L_{S}$) & 40.6 & 42.0 & 61.4 & 62.0 & 43.4 & 42.2 & 48.8 & 48.7 \\
    \noalign{\smallskip}
    \hline
    \noalign{\smallskip}
    w/o self-attention & 51.5 & 49.9 & 51.0 & 51.4 & 50.6 & 48.8 & 51.0 & 50.0 \\
    \noalign{\smallskip}
    \hline
    \noalign{\smallskip}
    w/o normals & 57.6 & 56.6 & 63.6 & 64.9 & 48.3 & 47.3 & 56.5 & 56.3 \\
    \hline
    \end{tabular}
}
\end{center}
\end{table}
\setlength{\tabcolsep}{1.4pt}

\subsection{Applications to shape analysis tasks}

As the pointwise features learned by the proposed unsupervised framework method are independent of the subsequent tasks, we demonstrate their applicability to the following point cloud analysis tasks: partial-to-partial shape registration, keypoint transfer via correspondence, and supervised part segmentation.

\subsubsection{Partial-to-partial shape registration.  } 
To perform partial shape registration between a shape and its rigid transform, we leverage the obtained pointwise correspondence between these two shapes, and compute a rigid transformation~\cite{arun1987least} to perform shape registration. To this end, we first pre-train our network on the pretext as described above. We then fine-tune it with more emphasis on the rigid transformation term $L_{R}$ by setting $\lambda_{C,R,S}=0.0001,1.0,0.06$. 

We compare, in a category-specific manner, our results to those produced by PRNet~\cite{wang2019prnet} on five categories of shapes from \textit{ModelNet40}~\cite{wu20153d}. 
We follow the training settings in \cite{wang2019prnet} by using 1024 points to train our network and PRNet with respect to each category data. 
It is worthy noting that different from their training strategy where a portion of points are subtracted from input data to mimic partial point clouds, we do not apply this specialized data augmentation to our training data. 
During test time, a shape with 1024 points is truncated to 768 points to produce a partial point cloud.
Both our method and PRNet perform 3 iterative estimations for pose registration.
Same random transformations are applied to generating a consistent testing set, ensuring a fair, category-specific comparison between our method and PRNet.

\begin{figure}[t!]
    \centering
    \includegraphics[width=\textwidth]{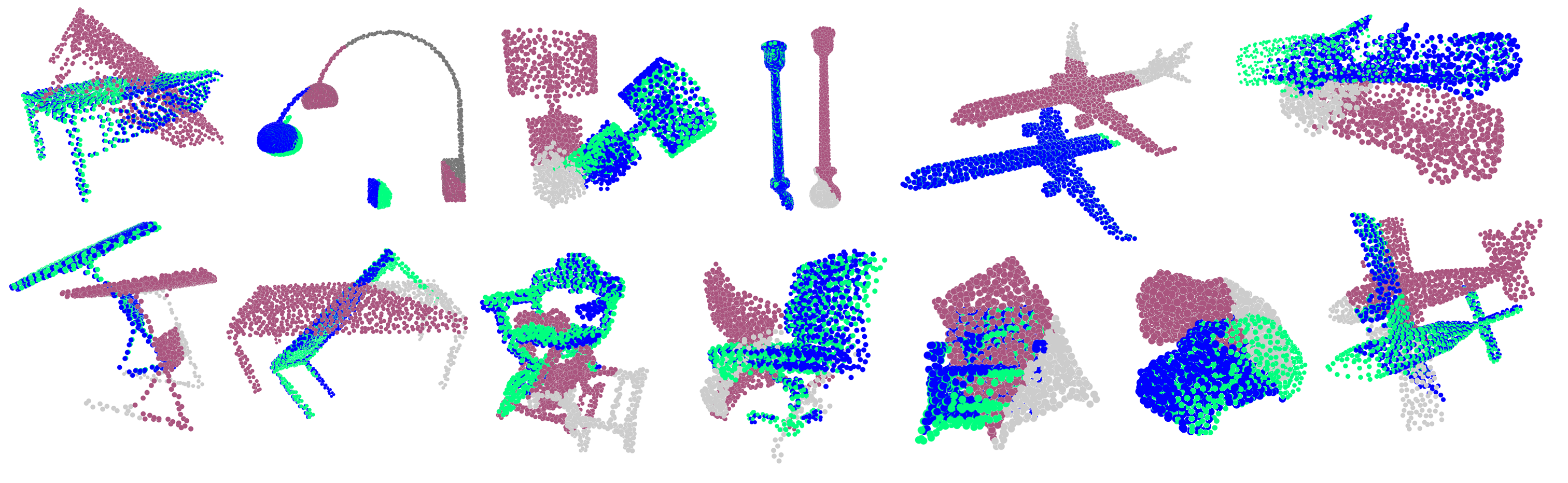}
    \caption{Visualization of partial-to-partial shape registration results. Purple and green point clouds are the source and target poses of a shape; the blue are the obtained results via three iterative match-and-transform; and the grey parts are those randomly truncated during the testing stage}
    \label{fig:registration}
\end{figure}

We evaluate the comparison results using the metrics Root Mean Square Error (RMSE) and Mean Absolute Error (MAE). 
As shown in Tab.~\ref{tab:registration_result}, our network can achieve results better than or at least comparable to PRNet (trained in a category-specific manner) across the listed categories. 
An exception is the \textit{Chair} category in terms of the rotation metrics.
Some registration results are randomly selected and visualized in Fig.~\ref{fig:registration} where shapes in purple, green and blue represent the source pose, target pose and our result, respectively.

\setlength{\tabcolsep}{4pt}
\begin{table}[t]
\begin{center}
\caption{Category-specific comparison with PRNet~\cite{wang2019prnet} for partial-to-partial registration on unseen point clouds. Bold values refer to the better performance.}
\label{tab:registration_result}
\scalebox{0.72}{
    \begin{tabular}{l| l || l l | l l | l l | l l | l l }
    \hline
    \noalign{\smallskip}
     \multicolumn{2}{c||}{Category} & \multicolumn{2}{c|}{Aeroplane} & \multicolumn{2}{c|}{Car} & \multicolumn{2}{c|}{Chair} & \multicolumn{2}{c|}{Table} & \multicolumn{2}{c}{Lamp} \\
    \noalign{\smallskip}
    \hline
    \noalign{\smallskip}
     \multicolumn{2}{c||}{Metric} & \cite{wang2019prnet} & Ours &\cite{wang2019prnet} & Ours &\cite{wang2019prnet} & Ours &\cite{wang2019prnet} & Ours
     &\cite{wang2019prnet} & Ours \\
    \noalign{\smallskip}
    \hline
    \noalign{\smallskip}
    \multirow{2}{*}{Rotation} 
     & RMSE & 7.206 & \textbf{4.287}  & 15.42 & \textbf{4.678} & \textbf{4.93} & 6.202 & 39.6 & \textbf{3.13} & 37.1 & \textbf{21.85} \\
     & MAE & 3.78 & \textbf{3.532} & 7.58 & \textbf{3.876} & \textbf{3.09} & 5.279 & 23.7 & \textbf{2.71} & 23.1 & \textbf{18.22} \\
    \noalign{\smallskip}
    \hline
    \noalign{\smallskip}
    \multirow{2}{*}{Translation} 
     & RMSE & 0.047 & \textbf{0.018} & 0.127 & \textbf{0.018} & 0.027 & \textbf{0.015} & 0.175 & \textbf{0.017} & 0.174 & \textbf{0.0476}  \\
     & MAE & 0.030 & \textbf{0.016} & 0.096 & \textbf{0.015} & 0.019 & \textbf{0.013} & 0.124 & \textbf{0.015} & 0.125 & \textbf{0.0418} \\
    \hline
    \end{tabular}
}
\end{center}
\end{table}
\setlength{\tabcolsep}{1.4pt}

\subsubsection{Keypoint transfer via correspondences.  }
Keypoints are a group of sparsely defined landmarks on a shape, crucial for many shape analysis applications. 
In this part, we demonstrate the learned pointwise representations can be leveraged to transfer a set of keypoints defined on a shape to other shapes alike. 
We compare our results to several state-of-the-art unsupervised learning methods for 3D shapes, i.e. \cite{deprelle2019learning,yang2018foldingnet} based on autoencoder structures and \cite{chen2019edgenet,muralikrishnan2019shape} based on pointwise feature learning (similar to ours). 
All methods are trained on the \textit{Airplane}, \textit{Chair}, and \textit{Bike} data from the \textit{ShapeNet part segmentation} dataset in a category-specifc, unsupervised manner. 
Shapes are pre-aligned for training and testing to conduct fair comparison. Evaluation is made on the test set in \cite{huang2017learning} where ground-truth keypoints of around 100 shapes per category are provided and each shape may contains 6 to 12 keypoints.

Given a point cloud shape, $\mathcal{P}$, with several 3D keypoints, we find their corresponding points in another given shape $\mathcal{Q}$. As the ground-truth keypoints are not given as a particular point in the point cloud data, we first sample 5 neighboring points from the point cloud and then search for each of the neighbors its correspondence point in $\mathcal{Q}$ via the learned pointwise features. Finally, we simply average the correspondence points to predict the corresponding keypoints in $\mathcal{Q}$.

\begin{figure}[t]
    \centering
    \includegraphics[width=0.9\textwidth]{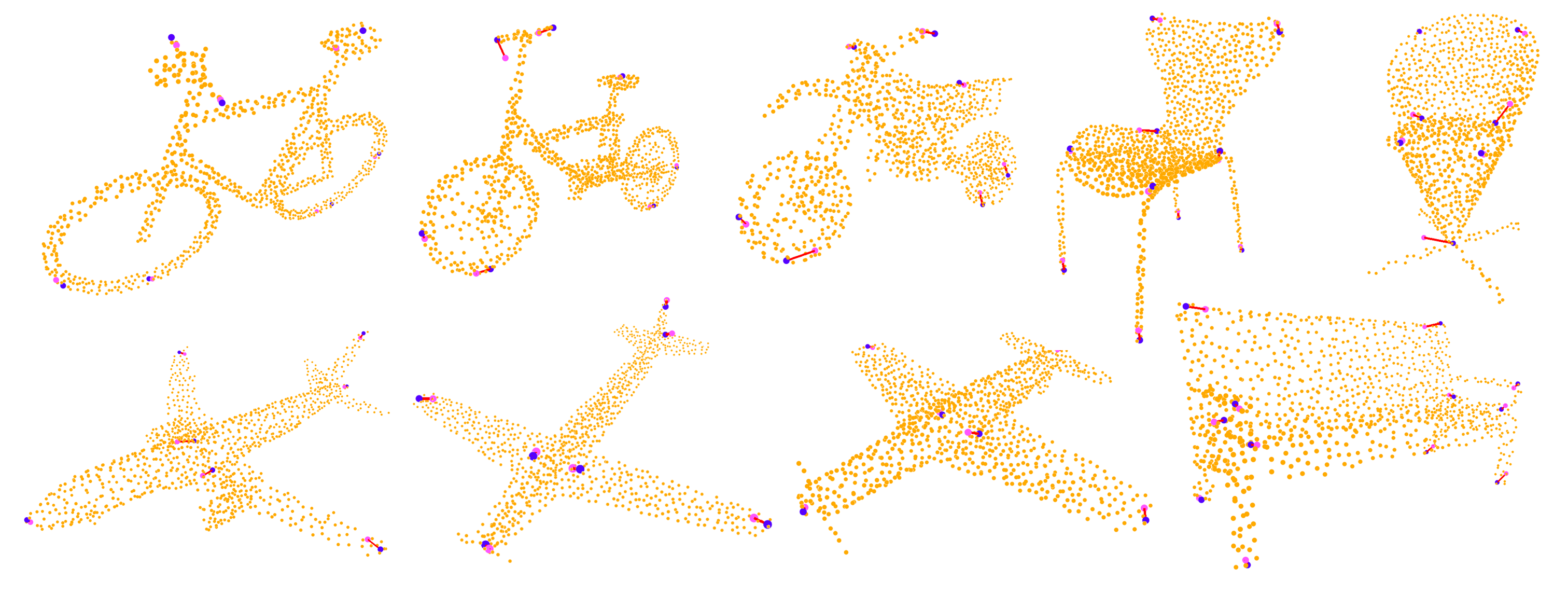}
    \caption{Visualization of the key point transfer task. Blue points are the ground-truth landmarks while points in magenta are estimated by our network}
    \label{fig:keypoint_experiments}
\end{figure}

We measure the distance error between the ground-truth keypoints and the predicted ones with the Euclidean metric. 
Noticing that the distance error greater than 0.05 is a relatively large value for a shape whose size is normalized with respect to its shape diameter, we show the percentage of keypoints with the distance error smaller than this threshold in Tab.~\ref{tab:keypoint_result}. 
As shown in the table, our result generally outperforms the other existing methods, except a slight fall behind ShapeUnicode~\cite{muralikrishnan2019shape} on the category of \textit{Bike}, showing the effectiveness of the learned pointwise representations in correspondence query. 
Some qualitative results are shown in Fig.~\ref{fig:keypoint_experiments}.

\setlength{\tabcolsep}{8pt}
\begin{table}[t]
\begin{center}
\caption{Results of keypoint transfer comparing with the state-of-the-art methods. The results are measured by the percentage of the keypoints with the distance error less than $0.05$. Bold values refer to the top performance}
\label{tab:keypoint_result}
\scalebox{0.72}{
    \begin{tabular}{c || cccccc}
    \hline
    \noalign{\smallskip}
     & LMVCNN\cite{huang2017learning} & AtlasNet\cite{deprelle2019learning} & FoldingNet\cite{yang2018foldingnet} & EdgeNet\cite{chen2019edgenet} & ShapeUnicode\cite{muralikrishnan2019shape} & Ours \\
     \noalign{\smallskip}
    \hline
    \noalign{\smallskip}
    Airplane & 30.3 & 51.1 & 26.6 & 33.5 & 30.8 & \textbf{57.9} \\
    Chair & 12.0  & 37.3 & 16.2 & 12.6 & 25.4 & \textbf{40.4} \\
    Bike& 17.4 & 34.2 & 31.7 & 27.2 & \textbf{58.3} & 49.8 \\
    \noalign{\smallskip}
    \hline
    \noalign{\smallskip}
    Mean  & 19.9 & 40.9 & 24.9 & 24.4 & 38.2 & \textbf{49.4} \\
    \hline
    \end{tabular}
}
\end{center}
\end{table}
\setlength{\tabcolsep}{1.4pt}

\subsubsection{Supervised part segmentation.  }
To demonstrate the ability of the learned pointwise representations as a feature to boost subsequent applications, we also use them as additional features to train a supervised part segmentation network. In particular, we adopt the PointNet++\cite{qi2017pointnet++} as a baseline for this supervised task. 

During this supervised training for part segmentation, the proposed network pre-trained on the dense correspondence pretext is frozen and serves as a feature extractor.
We compare the part segmentation results obtained with our additional input features and with the original inputs containing point coordinates and normals only.
As can be seen from Tab.~\ref{tab:segmentation}, the results with our pointwise features are improved in most categories. The part-averaged mean IoU (Intersection-over-Union) reaches $85.5\%$, higher than the performance obtained by the PointNet++.

On the other hand, when we train the network with $10\%$ of the labeled training data (without fine-tuning the pre-trained network), the performance gains are observed in 12 out of 16 categories, many of which outperforms the original results by a large margin (i.e. up to $7\%$). In this setting, we achieve a part-averaged mean IoU of $79.8\%$.

\setlength{\tabcolsep}{4pt}
\begin{table}[t]
\begin{center}
\caption{Comparison of the segmentation results on \textit{ShapeNet part segmentation} dataset trained with full and $10\%$ of the dataset}
\label{tab:segmentation}
\scalebox{0.58}{
    \begin{tabular}{c | l  l l l l l l l l l l l l l l l | l}
    \hline\noalign{\smallskip}
    Full train & aero. & bag & cap & car & chair & ear. & guitar & knife & lamp & laptop & motor & mug & pistol & rocket & skate. & table & mean \\
    \noalign{\smallskip}
    \hline
    \noalign{\smallskip}
    \cite{qi2017pointnet}	& 83.40 & 78.70 & 82.50 & 74.90 & 89.60 & \textbf{73.00} & 91.50 & 85.90 & 80.80 & 95.30 & 65.20 & 93.00 & 81.20 & 57.90 & 72.80 & 80.60 & 83.7 \\
    \cite{qi2017pointnet++} & 82.30 & 79.00 & \textbf{87.70} & 77.30 & \textbf{90.80} & 71.80 & 91.00 & 85.90 & \textbf{83.70} & 95.30 & 71.60 & 94.10 & 81.30 & 58.70 & \textbf{76.40} & 82.60 & 85.1 \\
    Ours & 82.66 & \textbf{81.97} & 79.96 & \textbf{78.03} & 85.77 & 70.12 & \textbf{91.61} & \textbf{86.53} & 81.81 & \textbf{96.03} & \textbf{73.55} & \textbf{95.57} & \textbf{83.49} & \textbf{59.10} & 75.39 & \textbf{88.23} & \textbf{85.5} \\
    \noalign{\smallskip}
    \hline
    \hline
    \noalign{\smallskip}
    $10\%$ train & aero & bag & cap & car & chair & ear. & guitar & knife & lamp & laptop & motor & mug & pistol & rocket & skate. & table & mean \\
    \noalign{\smallskip}
    \hline
    \noalign{\smallskip}
    \cite{qi2017pointnet}  & 76.10 & 69.80 & 62.60 & 61.40 & 86.00 & 62.10 & 86.20 & 79.70 & 73.60 & 93.30 & 59.10 & 83.40 & 75.90 & 41.80 & 57.70 & 74.80 & 77.3 \\
    \cite{qi2017pointnet++} & 76.40 & 43.40 & 77.80 & \textbf{75.52} & \textbf{87.50} & 67.70 & 87.40 & 77.40 & 71.40 & 94.10 & 61.30 & 90.40 & 72.80 & \textbf{51.40} & \textbf{68.70} & 75.30 & 78.6 \\
    Ours  & \textbf{77.09} & \textbf{73.24}  & \textbf{81.80} & 74.39 & 84.71 & \textbf{70.23} & \textbf{88.37} & \textbf{84.23} & \textbf{76.63} & \textbf{94.12} & \textbf{62.98} & \textbf{91.29} & \textbf{80.60} & 51.25 & 65.02 & \textbf{77.94} & \textbf{79.8} \\
    \hline
    \end{tabular}
}
\end{center}
\end{table}

\setlength{\tabcolsep}{1.4pt}

\section{Conclusions and Future Work}

This paper proposes a pretext of finding dense correspondences between two different shapes for unsupervised learning of pointwise features for point cloud shapes and formulates a cycle-consistency based framework to solve this problem. 
In order to learn discriminative pointwise features, we force the cycle correspondences to be as bijective as possible using the Sinkhorn regularization. 
Ablation study validates the design and effectiveness of the proposed unsupervised framework. Furthermore, we demonstrate the applicability of acquired pointwise features in downstream tasks: partial-to-partial shape registration, unsupervised keypoint transfer, and supervised part segmentation.

While geometric correspondences can be effectively learned by the proposed approach, such correspondences learned in this unsupervised manner may fail to capture the semantic meaning of the shapes. As future work, we would like to explore solutions to this problem based on the proposed unsupervised framework.

\section*{Acknowledgement}
We acknowledge the valuable comments from anonymous reviewers. This work is supported in part by ITF Grant ITS/457/17FP.

\bibliographystyle{splncs04}
\bibliography{egbib}

\clearpage
\section*{Appendix}

More qualitative results and details regarding the network architecture are provided in this supplementary.

\subsection*{A.1 Qualitative results of dense correspondences between shapes}
\begin{figure}
    \centering
    \includegraphics[width=\textwidth]{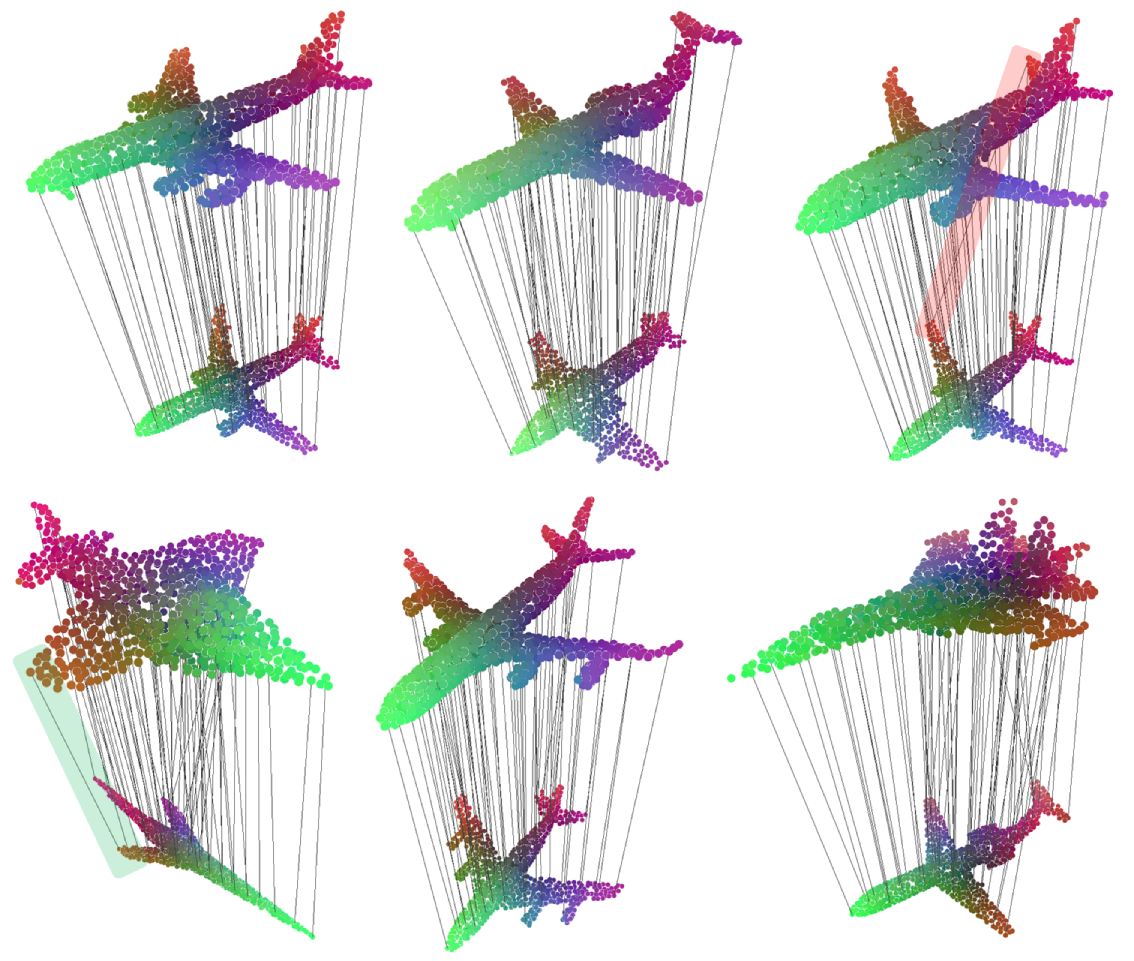}
    \caption{Dense correspondences between different shapes from the Airplane category. Highly similar local structures (e.g. the wing tip and the stabilizer tip at tail highlighted by the red block in top-right) lead to undesired correspondences. Even the shapes are quite different from each other in bottom left, correct correspondences (highlighted in green) are obtained, showing the robustness of the proposed method in handling inter-instance variation}
    \label{fig:dense_corr_airplane}
\end{figure}

\begin{figure}
    \centering
    \includegraphics[width=\textwidth]{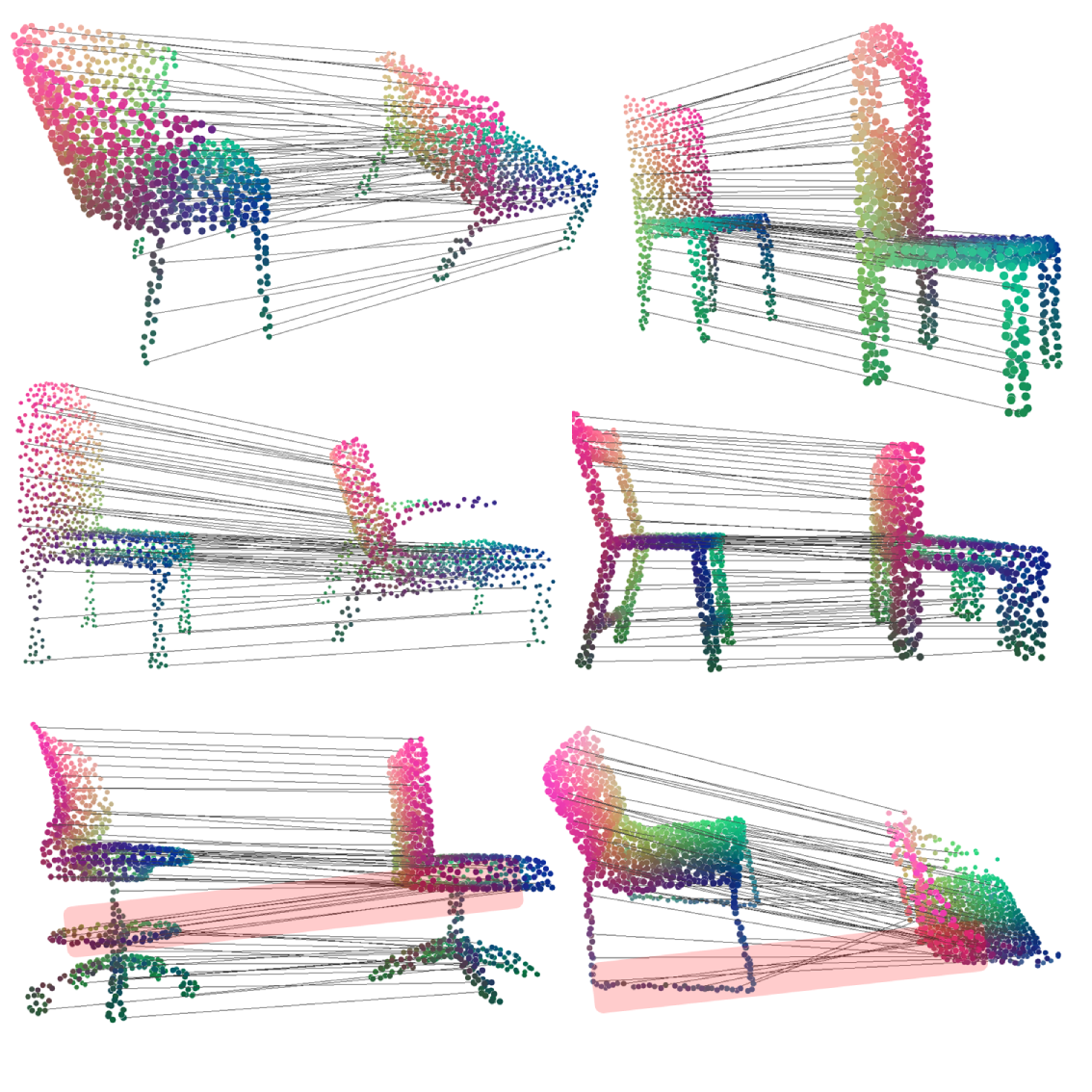}
    \caption{Dense correspondences between different shapes from the Chair category. Inconsistent topological structures (see the two examples at the bottom row) lead to non-intuitive correspondences.}
    \label{fig:dense_corr_chair}
\end{figure}

\begin{figure}
    \centering
    \includegraphics[width=\textwidth]{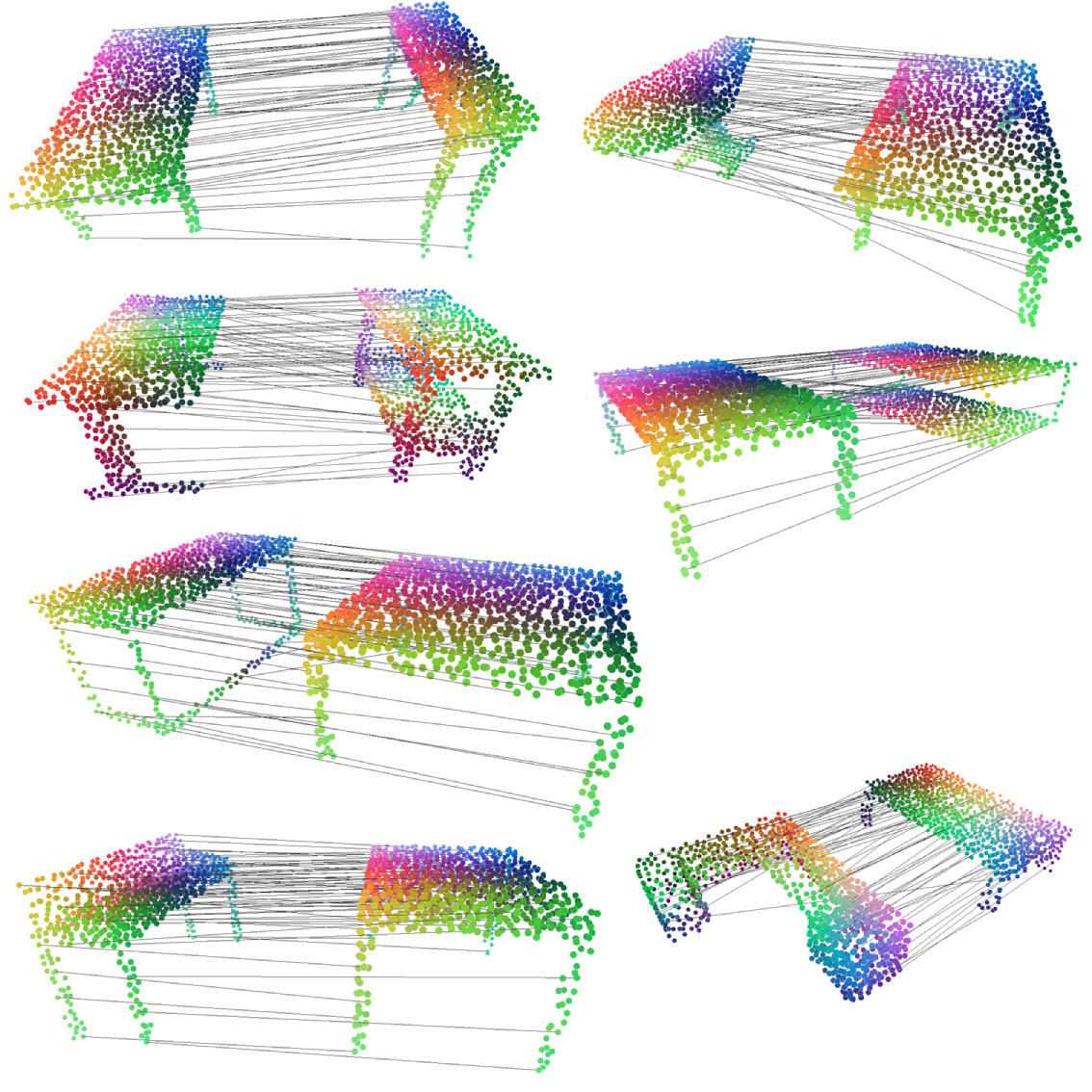}
    \caption{Dense correspondences between different shapes from the Table category}
    \label{fig:dense_corr_table}
\end{figure}

We randomly sample some pairs of shapes from different categories and visualize the obtained dense correspondences between each pair of shapes in Figs.~\ref{fig:dense_corr_airplane},~\ref{fig:dense_corr_chair}, and~\ref{fig:dense_corr_table}.
For better visualization, we color the point set by reducing the dimension of the pointwise representations to 3 using t-SNE~\cite{maaten2008visualizing}. Only a sparse set of correspondence links are sampled and visualized. Some of the undesired correspondence links are highlighted by overlaying red blocks on them. 
These undesired correspondences are mainly due to: 
1) highly similar local structures (see the airplane example where the wing tip and the stablizer tip at tail are linked);
2) inconsistent topological structures (see the two examples in chairs).

\subsection*{A.2 Network architecture}

\textbf{Point-based network. }
The PointNet++ architecture for our pre-trained network consists of three set abstraction (SA) layers, three feature propagation (FP) layers, and a fully-connected layer at the end. 
The multi-scale grouping strategy is used in the adopted network. An input point cloud with 1024 points is fed to the PointNet++ backbone. 
In the experiments of Keypoint transfer and Supervised part segmentation, point normals are used as input to the network.
The PointNet++ then processes the point cloud to produce a pointwise 128-dimensional features as output. 
Finally, the FC layer reduces the 128-dimensional output from the PointNet++ backbone to a 64-dimensional output, yielding the final pointwise representaitons. The details are listed in Tab.~\ref{tab:architecture}:

\setlength{\tabcolsep}{4pt}
\begin{table}[h]
\begin{center}
\caption{Network architecture}
\label{tab:architecture}
\scalebox{0.94}{
    \begin{tabular}{ l | l | l | l }
    \hline
    \noalign{\smallskip}
    Layer & Channel sizes of MLPs & $\#$ Input points & Multi-scale radii \\
    \hline
    SA1 & $[32,32,64],[64,64,128],[64,96,128]$ & 512 & $[0.1,0.2,0.4]$ \\
    \hline
    SA2 & $[128,128,256],[128,196,256]$ & 256 & $[0.4,0.8]$ \\
    \hline
    SA3 & $[256,512,1024]$ & n.a. & n.a. \\
    \hline
    FP3 & $[512,256]$ & n.a. & n.a. \\
    \hline
    FP2 & $[256,256]$ & n.a. & n.a. \\
    \hline
    FP1 & $[128,128]$ & n.a. & n.a. \\
    \hline
    FC & $[128,64]$ & n.a. & n.a. \\
    \hline
    \end{tabular}
}
\end{center}
\end{table}
\setlength{\tabcolsep}{1.4pt}

\noindent\textbf{Self-attention module. }
The set attention blocks proposed (Equation 8 in \cite{lee2018set}) are adopted as a self-attention module and inserted in-between each adjacent layers listed in Tab.~\ref{tab:architecture}. We instantiate this module with 4 heads and the LayerNorm operation.

\end{document}